\documentclass[letterpaper]{article} 

\usepackage{aaai2027}
\nocopyright

\usepackage[hyphens]{url}  
\usepackage{graphicx} 
\usepackage{natbib}  
\usepackage{caption} 
\usepackage{algorithm}
\usepackage{algorithmic}
\usepackage{newfloat}
\usepackage{listings}
\DeclareCaptionStyle{ruled}{labelfont=normalfont,labelsep=colon,strut=off} 
\floatstyle{ruled}
\newfloat{listing}{tb}{lst}{}
\floatname{listing}{Listing}
\usepackage{booktabs}
\usepackage{amsmath}
\usepackage{amssymb}
\usepackage{multirow}
\usepackage{float}

\title{MedLLM: An Open Medical Language Model at the Sub-Billion Scale}

\author {
    Maxx Richard Rahman\textsuperscript{\rm 1,\rm 2}\corresponding,
    Asim Ahmed\textsuperscript{\rm 1},
    Mihan Mohagheghzadeh\textsuperscript{\rm 1},
    Wolfgang Maass\textsuperscript{\rm 1,\rm 2}
}
\affiliations {
    \textsuperscript{\rm 1}German Research Center for Artificial Intelligence (DFKI), Germany\\
    \textsuperscript{\rm 2}Saarland University, Germany\\
    maxx\_richard.rahman@dfki.de, 
    asim.ahmed@dfki.de,
    mihan.mohagheghzadeh@dfki.de,
    wolfgang.maass@dfki.de
}

\begin{document}

\maketitle

\begin{abstract}
Open medical language models have converged on a single scale: every
widely used system runs at 7B parameters or more, leaving the
sub-billion regime uncharacterized. We present MedLLM, an open
0.1B-parameter medical language model trained through a fully open
three-phase pipeline: general pretraining with curriculum
sequence-length scheduling, domain fine-tuning on MedFineWeb, a
reference-guided medical corpus we release that is selected from general
web data by embedding similarity to medical question-answering (QA)
data, and preference-aligned fine-tuning combining SFT with direct
preference optimization (DPO). Across medical benchmarks, MedLLM shows a
pattern visible only at sub-billion scale: medical competence does not
degrade uniformly under compression but splits by task type. On
context-grounded QA it comes within $2.9$~pp of a medically adapted 7B
model and surpasses the instruction-tuned and general-purpose 7B
baselines; on knowledge-recall QA it stays near the task floor on
clinical-vignette MedQA yet significantly exceeds every 7B and sub-7B
baseline on MedMCQA, indicating that where recall fails the constraint is
model capacity rather than adaptation. This dissociation is masked at 7B,
where both capabilities are present, and surfaces only when capacity is
scarce.
\end{abstract}

\section{Introduction}

Evidence-based medicine relies on the accurate recall, synthesis, and
application of vast bodies of clinical and biomedical
knowledge~\cite{sackett1996evidence}. Large language models (LLMs)
encode and retrieve such knowledge at
scale~\cite{singhal2023large,nori2023capabilities,brown2020language},
prompting substantial interest in medical LLMs for clinical
decision-making, medical education, and biomedical
research~\cite{thirunavukarasu2023large}. Yet the most capable medical
LLMs are either closed-source with undisclosed training details, such
as GPT-4 and Med-PaLM-2~\cite{nori2023capabilities,singhal2023towards},
or open but large, operating at 7B to 540B
parameters~\cite{chen2023meditron,singhal2023large,peng2023study}.
Both properties limit access and reproducibility, and they leave a
basic question unanswered: below 7B, which medical capabilities
survive, and which collapse? The entire sub-billion regime, where a
model is small enough to train, audit, and deploy on modest hardware,
remains uncharacterized.
 
Adapting an LLM to medicine is difficult because of distributional
mismatch. General-purpose models are trained on web text whose
distribution $P_{\text{web}}$ diverges sharply from medical text
$P_{\text{med}}$~\cite{gururangan2020dont,lee2020biobert}: medical
language combines long-tail terminology, structured clinical
reasoning, domain-specific abbreviations, and high-precision
requirements where factual errors carry serious
consequences~\cite{gu2021domain,peng2023study}. Prior work narrows
this gap through continued pretraining on biomedical
corpora~\cite{wu2023pmcllama,chen2023meditron}, instruction tuning on
medical QA~\cite{han2023medalpaca,zhang2024alpacare}, and preference
optimization~\cite{singhal2023towards}, and continued pretraining is
known to allocate a model's capacity toward the target distribution
rather than the full breadth of
$P_{\text{web}}$~\cite{gururangan2020dont,gupta2023continual}. Every
one of these efforts, however, operates at 7B parameters or more. As a
result, it is unknown whether the same recipe transfers to a model an
order of magnitude smaller, or whether reduced capacity forces some
medical capabilities to break before others. Key contributions:

\begin{enumerate}
    \item We propose MedLLM, a 0.1B-parameter open medical language
    model, with its full three-phase pipeline, placing a reproducible medical modeling baseline an order of magnitude below existing open medical LLMs.

    \item We introduce MedFineWeb, a medical pretraining corpus selected
    from general web text by embedding similarity to medical QA, which
    requires no curated medical source and transfers to any domain given
    a small set of target-domain QA examples.

    \item We identify a task-structured dissociation under compression: 
    the 0.1B model reaches 7B-competitive
    accuracy on context-grounded QA yet remains capacity-bound on
    knowledge-recall QA, as the binding constraint at sub-billion size.
\end{enumerate}

\section{Related Works}
\subsection{Medical Language Models and Domain Adaptation}
Continued pretraining on domain-specific corpora is the dominant
paradigm for building medical LLMs. Early encoder-only efforts,
BioBERT~\cite{lee2020biobert} and PubMedBERT~\cite{gu2021domain},
pretrained BERT on PubMed text and improved biomedical NLP, while
\citet{gururangan2020dont} showed that domain- and task-adaptive
pretraining yield additive gains, confirmed at larger
scales~\cite{gupta2023continual,ke2023continual}. With decoder-only
architectures the approach scaled sharply: PMC-LLaMA~\cite{wu2023pmcllama}
continued pretraining LLaMA-7B on 4.8M PubMed Central papers;
Meditron~\cite{chen2023meditron} extended this to 7B and 70B parameters
on a curated 48.1B-token corpus of PubMed articles, abstracts, and
clinical guidelines, establishing open baselines that approach
closed-source systems~\cite{singhal2023large,singhal2023towards};
GatorTronGPT~\cite{peng2023study} pretrained a 20B model on 227B
clinical and English tokens; MedAlpaca~\cite{han2023medalpaca} and
AlpaCare~\cite{zhang2024alpacare} instruction-tuned at 7B and 13B;
Med-PaLM and Med-PaLM~2~\cite{singhal2023large,singhal2023towards}
instruction-tuned 540B-parameter models to near expert-level
licensing-exam performance; and Clinical-Camel~\cite{toma2023clinical}
adapted Llama-2-70B via QLoRA on clinical dialogues. All operate at 7B
parameters or above and depend on large, manually curated medical
sources. Medical domain adaptation at substantially smaller scale, and
from web-selected rather than curated data, remains unexplored, the gap
MedLLM and MedFineWeb address.

\subsection{Preference Optimization for Language Model Alignment}
Reinforcement learning from human feedback (RLHF) trains a separate
reward model from preference data and optimizes the policy via proximal
policy optimization~\cite{ouyang2022training,christiano2017deep}, while
direct preference optimization (DPO)~\cite{rafailov2024direct} removes
the reward model by reparametrizing the reward in terms of the policy
itself, provably optimizing the same objective under the Bradley-Terry
model~\cite{bradley1952rank}. DPO has been applied broadly, for instance
Zephyr~\cite{tunstall2023zephyr}, which used distilled DPO to make a 7B
model competitive with larger instruction-tuned systems, and in medicine
HuatuoGPT-II~\cite{chen2024huatuogpt}, which applied one-stage
preference alignment, alongside RLHF efforts aimed at reducing
hallucination in clinical text~\cite{thirunavukarasu2023large}. Its
behavior on small-scale medical models, however, has not been studied.
We provide the first analysis of DPO for medical QA at the 0.1B scale,
showing that preference alignment sharply amplifies the log-probability
margin between correct and incorrect answers while yielding only limited
accuracy change, evidence that at this scale DPO refines existing
discrimination rather than creating new knowledge.

\section{MedLLM Model}

\begin{figure*}[t]
\centering
\includegraphics[width=0.9\textwidth]{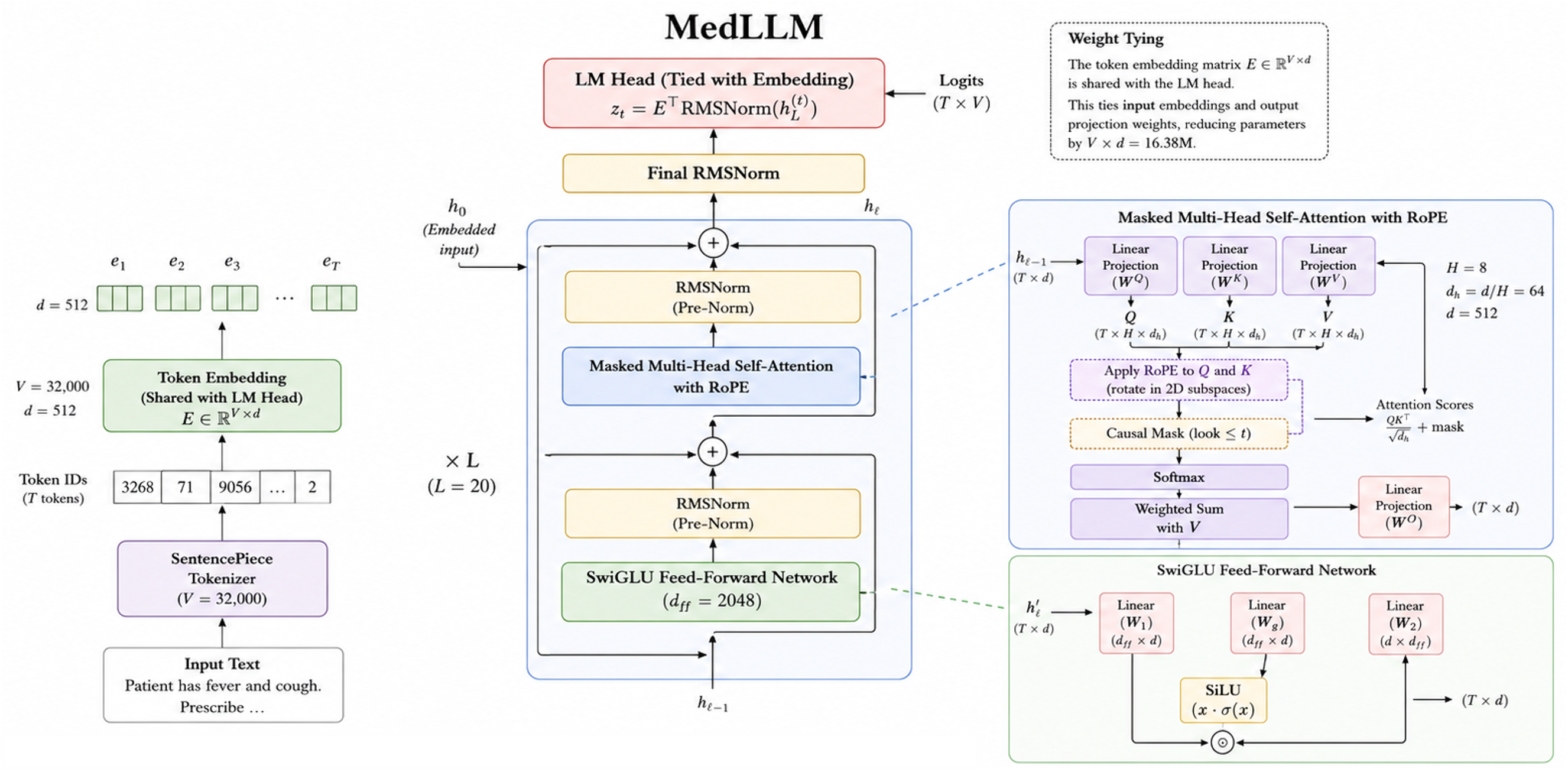}
\caption{MedLLM model: 0.1B-parameter decoder-only transformer with 20 layers.}
\label{fig:overview}
\end{figure*}

\subsection{Model Architecture}
\subsubsection{Tokenization and Embedding.}
MedLLM is a decoder-only transformer of 0.1B parameters
(Figure~\ref{fig:overview}). Inputs are tokenized with a SentencePiece~\cite{kudo2018sentencepiece}
vocabulary of $V = 32{,}000$ and embedded into $d = 512$ dimensions,
mapping each token $x_t$ to $\mathbf{e}_t = \mathbf{E}_{x_t}$ with
$\mathbf{E} \in \mathbb{R}^{V \times d}$. The vocabulary-to-width ratio
$V/d = 62.5$ is large, so an untied output projection would be the
single largest tensor in the model, exceeding any transformer layer. We
therefore tie the input embedding to the output
projection~\cite{press2017using}: tying removes a separate
$V \times d = 16.38$M-parameter matrix, $14\%$ of the untied
$116.7$M-parameter total, and constrains the input and output
representations of each token to share one geometry, which reduces the
effective parameter count and acts as a regularizer that is known to help
most when $d$ is small relative to $V$.

\subsubsection{Transformer Layers.}
The model stacks $L = 20$ layers. Each layer places
RMSNorm on the \emph{output} of the attention and feed-forward sublayers
rather than on their input, so normalization sits inside the residual
branch:
\begin{align}
    \mathbf{h}_\ell' &= \mathbf{h}_{\ell-1} + \text{RMSNorm}(\text{MHA}(\mathbf{h}_{\ell-1})), \label{eq:attn_block} \\
    \mathbf{h}_\ell &= \mathbf{h}_\ell' + \text{RMSNorm}(\text{SwiGLU}(\mathbf{h}_\ell')), \label{eq:ffn_block}
\end{align}
with $\mathbf{h}_0$ the embedded sequence. This reordered placement
normalizes each sublayer's contribution before it is added back, which
keeps the residual stream at a controlled scale across depth and
improves training stability over the input-normalized pre-norm
arrangement~\cite{walsh2025olmo2}. We use RMSNorm~\cite{zhang2019root},
\begin{equation}\label{eq:rmsnorm}
    \text{RMSNorm}(\mathbf{x}) = \frac{\mathbf{x}}{\text{RMS}(\mathbf{x})} \odot \mathbf{g},
    \qquad
    \text{RMS}(\mathbf{x}) = \sqrt{\tfrac{1}{d}\textstyle\sum_{i=1}^d x_i^2},
\end{equation}
which drops the mean-centering of LayerNorm and enforces scale
invariance of its argument, lowering per-step cost while preserving
optimization stability. Because every sublayer remains a perturbation of
the identity, gradients propagate through the depth-20 stack without the
vanishing that would otherwise threaten a deep, narrow model.

\subsubsection{Masked Multi-Head Self-Attention with RoPE.}
Attention uses $H = 8$ heads of dimension $d_h = d/H = 64$, with
per-head projections
$\mathbf{Q}_i = \mathbf{h}\mathbf{W}_i^Q$,
$\mathbf{K}_i = \mathbf{h}\mathbf{W}_i^K$,
$\mathbf{V}_i = \mathbf{h}\mathbf{W}_i^V$ and
$\mathbf{W}_i^Q, \mathbf{W}_i^K, \mathbf{W}_i^V \in \mathbb{R}^{d \times d_h}$.
Queries and keys are RMSNorm-normalized per head before
the dot product (QK-norm), which bounds attention-logit magnitudes and
prevents the score blow-ups that destabilize small-model training.
Relative position is supplied by rotary embeddings
(RoPE)~\cite{su2024roformer}, which apply to the query at position $m$
and the key at position $n$ a block-diagonal rotation
$\mathbf{R}_{\Theta,m}$ composed of $2$D rotations by angle $m\theta_j$
in the $j$-th coordinate pair, with frequencies
$\theta_j = b^{-2j/d_h}$ and base $b = 10{,}000$. Because rotations
compose additively,
$\mathbf{R}_{\Theta,m}^\top \mathbf{R}_{\Theta,n} = \mathbf{R}_{\Theta,\,n-m}$,
the attention score depends on the two positions only through their
offset,
\begin{equation}\label{eq:rope_relative}
    (\mathbf{R}_{\Theta,m}\mathbf{q})^\top (\mathbf{R}_{\Theta,n}\mathbf{k})
    = \mathbf{q}^\top \mathbf{R}_{\Theta,\,n-m}\,\mathbf{k},
\end{equation}
so relative position is encoded multiplicatively inside the score
without additive position embeddings and without consuming parameters. A
causal mask restricts token $t$ to positions $\leq t$.

\subsubsection{SwiGLU Feed-Forward Network.}
The feed-forward block uses SwiGLU~\cite{shazeer2020glu}:
$
    \text{SwiGLU}(\mathbf{h}) = (\mathbf{W}_1 \mathbf{h} \odot \text{SiLU}(\mathbf{W}_g \mathbf{h})) \mathbf{W}_2,
$
with $\mathbf{W}_1, \mathbf{W}_g \in \mathbb{R}^{d_{\text{ff}} \times d}$,
$\mathbf{W}_2 \in \mathbb{R}^{d \times d_{\text{ff}}}$,
$d_{\text{ff}} = 2048 = 4d$, and $\text{SiLU}(x) = x \cdot \sigma(x)$. The
data-dependent gate $\text{SiLU}(\mathbf{W}_g\mathbf{h})$ modulates each
hidden unit multiplicatively, which improves language-modeling
perplexity over the additive nonlinearities of ReLU and GELU
gating~\cite{shazeer2020glu}. The gate adds a third projection, so the
block costs $3\,d\,d_{\text{ff}}$ parameters against $2\,d\,d_{\text{ff}}$
for a standard two-matrix feed-forward network of equal width; at
$d_{\text{ff}}=4d$ this places $62.7\%$ of the model's parameters in
position-wise feature transformation (Table~\ref{tab:budget}), a
deliberate allocation toward the sublayer where, at small scale, added
capacity is used most efficiently.

\subsubsection{Output Projection.}
A final RMSNorm precedes the tied output projection,
$
    \mathbf{z}_t = \mathbf{E}^\top \text{RMSNorm}(\mathbf{h}_L^{(t)}) \in \mathbb{R}^V,
$
producing the vocabulary logits at each position. These
logits are additionally regularized during training by a z-loss term
that penalizes their log-partition function.

\subsubsection{Parameter Allocation.}
Table~\ref{tab:budget} accounts for the full budget. With
$V/d = 62.5$, the tied embedding already consumes $16.3\%$ of the model,
so of the $\sim\!84$M parameters that remain for computation the design
places the majority in the feed-forward sublayers ($62.7\%$) and the
remainder in attention ($20.9\%$), spending depth ($L=20$) rather than
width to favor compositional transformation over a wider, shallower
alternative at equal budget, consistent with evidence that at fixed
parameter count depth contributes more than width until it
saturates~\cite{levine2020limits}. This accounting makes the central
constraint of the paper concrete: at $0.1$B parameters there is no
capacity to spare for redundant tensors.

\begin{table}[t]
\centering
\caption{Parameter allocation of MedLLM
($d{=}512$, $L{=}20$, $H{=}8$, $d_{\text{ff}}{=}2048$, $V{=}32{,}000$).}
\label{tab:budget}
\begin{tabular}{lrr}
\toprule
\textbf{Component} & \textbf{Parameters} & \textbf{Share} \\
\midrule
Token embedding (tied)      & 16.38M  & 16.3\% \\
Attention ($20\times$)      & 20.97M  & 20.9\% \\
Feed-forward ($20\times$)   & 62.91M  & 62.7\% \\
Normalization gains         & 0.02M   & $<0.1\%$ \\
\midrule
Total                       & 100.29M & 100\% \\
\bottomrule
\end{tabular}
\end{table}

\subsection{Reference-Guided Medical Corpus}

MedLLM is domain-adapted on MedFineWeb, a medical plain-text corpus we
construct from general web data by reference-guided selection.

\subsubsection{General Pretraining Corpus.}
The general corpus $\mathcal{D}_{\text{gen}}$ comprises 46{,}083{,}165
documents totaling approximately 29.5B tokens (45~GB), drawn from
RedPajama-V2~\cite{together2023redpajama} (7{,}554{,}494 documents),
C4~\cite{raffel2020exploring} (32{,}000{,}000 documents),
OpenWebText~\cite{gokaslan2019openwebtext} (8{,}013{,}769 documents),
and Wikipedia (6{,}407{,}814 documents). Tokenized into 29{,}498 shards
at 634.13 tokens per document on average, it provides the broad
linguistic coverage on which general pretraining is performed and from
which MedFineWeb is selected.

\subsubsection{Candidate Pool and Reference Sets.}
PubMed supplies abundant biomedical text but is dominated by
abstract-style scientific writing, which underrepresents the
explanatory, health-information, and question-oriented style of
downstream medical QA. To cover that style without a manually curated
medical source, we select medical passages directly from general web
text. We split $\mathcal{D}_{\text{gen}}$ into fixed-size candidate
chunks of approximately 1000 words, yielding 2M candidates, each
retaining source metadata, and score every candidate by semantic
similarity to medical QA data.

We use three medical QA datasets as semantic references only: MedMCQA,
MedQA, and PubMedQA~\cite{pal2022medmcqa,jin2021disease,jin2019pubmedqa}.
These datasets are never added to MedFineWeb as text; they define what
medical content looks like for retrieval. For each
reference example we concatenate its prompt and response fields, and to
keep retrieval tractable we use up to 20{,}000 references per dataset
(20{,}000 MedMCQA, 10{,}178 MedQA, and 20{,}000 PubMedQA).

\paragraph{Similarity Scoring.}
All candidate chunks and reference examples are embedded with the same
Sentence Transformer~\cite{reimers2019sentencebert} and L2-normalized,
so that inner product equals cosine similarity. For each reference
dataset $R$ we build a FAISS inner-product
index~\cite{johnson2019billion}, which compares 2M candidates against
thousands of references without explicit pairwise computation. For each
candidate chunk $c$ we retrieve its top-5 nearest references and score it
by their average cosine similarity,
$
s_R(c) = \frac{1}{5}\sum_{i=1}^{5}\cos\!\left(f_\phi(c), f_\phi(r_i)\right),
$
where $f_\phi$ is the embedding model and $r_i$ is the $i$-th nearest
reference in $R$. Averaging over five neighbors rather than taking the
single nearest makes the score robust to an accidental match and
measures similarity to a local neighborhood of medical content rather
than to one reference point.

\begin{algorithm}[tb]
\caption{Reference-Guided MedFineWeb Construction}
\label{alg:medfineweb_construction}
\textbf{Input}: General-domain collection $\mathcal{D}_{\text{gen}}$; medical QA reference sets $\mathcal{R}=\{\text{MedMCQA}, \text{MedQA}, \text{PubMedQA}\}$; embedding model $f_\phi$; top fraction $\alpha=0.50$\\
\textbf{Output}: MedFineWeb plain-text corpus
\begin{algorithmic}[1]
\STATE Split $\mathcal{D}_{\text{gen}}$ into 1000-word chunks $\{c_1,\ldots,c_N\}$
\FOR{each reference set $R \in \mathcal{R}$}
    \STATE Concatenate prompt and response of each reference example
    \STATE Embed up to 20000 references with $f_\phi$; build a FAISS inner-product index
    \FOR{each candidate chunk $c_i$}
        \STATE Embed $c_i$; retrieve top-5 nearest references
        \STATE $s_R(c_i) \leftarrow$ average top-5 cosine similarity
    \ENDFOR
    \STATE Rank candidates by $s_R(c_i)$ and keep the top $\alpha N$
\ENDFOR
\STATE Merge the per-reference selections and deduplicate at the text level
\RETURN MedFineWeb
\end{algorithmic}
\end{algorithm}

\paragraph{MedFineWeb.}
For each reference set we rank all 2M candidates by $s_R$ and keep the
top 50\%, giving three per-reference selections of 1M chunks each
(Table~\ref{tab:medfineweb}). Although the three rankings draw from a
shared candidate pool, they select largely disjoint halves: the top-50\%
sets by similarity to MedMCQA, MedQA, and PubMedQA overlap only
marginally, so merging and text-level deduplication remove a negligible
fraction and yield MedFineWeb at 3M chunks. This near-disjointness is
itself informative, indicating that the three reference styles
(multiple-choice exam questions, USMLE vignettes, and abstract-based
research questions) retrieve different regions of the web corpus rather
than converging on one medical subset. MedFineWeb totals approximately
3B SentencePiece tokens, averaging about 1000 tokens per chunk, remains
entirely plain text, and is never converted to QA format.

\begin{table}[ht]
\centering
\caption{Per-reference selections comprising MedFineWeb (top 50\% by
similarity to each reference set).}
\label{tab:medfineweb}
\begin{tabular}{lrr}
\toprule
\textbf{Reference Set} & \textbf{Selected Chunks} & \textbf{Size} \\
\midrule
MedMCQA top 50\%  & 1M & 6.48GB \\
MedQA top 50\%    & 1M & 6.45GB \\
PubMedQA top 50\% & 1M & 6.48GB \\
\bottomrule
\end{tabular}
\end{table}

\subsubsection{Leakage and Contamination Analysis.}
Because the reference sets also serve as evaluation benchmarks, we verify
that MedFineWeb contains no benchmark examples. Its chunks are drawn only
from $\mathcal{D}_{\text{gen}}$, never from the benchmark data. Comparing
the selected chunks against MedMCQA, MedQA, and PubMedQA by normalized
full-text hashing finds no exact matches, and a 20-word $n$-gram overlap
check flags 22 chunks (about 0.0018\%), which manual inspection confirms
are shared biomedical phrasings rather than benchmark items. However, verbatim checks do not address the deeper concern that
selection by similarity to the benchmarks could inject answer knowledge
even without copying text. Our results indicate it does not, by an
argument internal to the outcomes. Were similarity-based selection
leaking answers, the datasets used as references would be the ones to
improve, and would improve together; instead the effect is
task-structured, not reference-structured: MedMCQA and MedQA, both
references, remain near the task floor after adaptation, while the gain
concentrates on context-grounded PubMedQA, where the answer is
recoverable from the provided passage rather than from selected
pretraining text. A leakage account predicts the opposite pattern, so
the evidence runs against answer-level contamination rather than for it.
We therefore read the results as in-domain medical performance under
reference-guided adaptation: selection shifts MedFineWeb toward medical
topic and style without supplying benchmark answers.

\subsection{Training and Optimization}
MedLLM training proceeds in two stages: pretraining, comprising general
pretraining and domain fine-tuning on MedFineWeb, followed by
task-specific alignment via supervised fine-tuning (SFT) and direct
preference optimization (DPO). All configurations are given in
Table~\ref{tab:sft_config}.

\subsubsection{General Pretraining.}
From random initialization, we pretrain the 0.1B model on
$\mathcal{D}_{\text{gen}}$ (approximately 29.5B tokens) for 200{,}000
steps on 2$\times$ NVIDIA H100 GPUs under the causal language modeling
objective
\begin{equation}\label{eq:clm}
\mathcal{L}_{\text{CLM}}(\theta) = -\sum_{t} \log p_\theta(x_t \mid x_{<t}),
\end{equation}
to which we add the z-loss regularizer on the output logits,
\begin{equation}\label{eq:zloss}
\mathcal{L}(\theta) = \mathcal{L}_{\text{CLM}}(\theta)
+ \lambda_z\, \mathbb{E}_t\!\left[\log^2 \textstyle\sum_{v} \exp(z_{t,v})\right],
\end{equation}
where $z_{t,v}$ is the logit for vocabulary item $v$ at position $t$ and
$\lambda_z = 10^{-4}$. The z-loss penalizes drift in the logit
log-partition function, keeping the softmax normalizer near unity and
preventing the output-scale instabilities that a tied embedding of
$V=32{,}000$ can otherwise induce at small width. We apply curriculum
sequence-length scheduling~\cite{bengio2009curriculum}, raising the
maximum length in three stages ($128 \to 256 \to 512$ tokens), so the
model consolidates local dependencies before attending to longer
contexts, which stabilizes optimization at small scale. The final
checkpoint reaches a validation loss of 4.005, with
loss falling rapidly early in training and improving steadily
thereafter.

\subsubsection{Domain Fine-Tuning (DFT).}
We continue pretraining on MedFineWeb under the same objective for
25{,}000 steps at a learning rate of $5 \times 10^{-5}$, adapting the
model from the general distribution $P_{\text{web}}$ toward medical text
$P_{\text{med}}$. Validation perplexity decreases monotonically, from
27.3 at step 1{,}000 to 15.68 at step 8{,}000 and 13.62 at step
25{,}000, with no overfitting; we carry the final step-25{,}000
checkpoint into all subsequent phases.

\subsubsection{Supervised Fine-Tuning (SFT).}
To measure per-benchmark generalization, we fine-tune a separate SFT
model for each benchmark from the shared DFT checkpoint, at learning rate
$2 \times 10^{-5}$ under the label-prediction objective
\begin{equation}\label{eq:sft}
\mathcal{L}_{\text{SFT}}(\theta) = -\log p_\theta(y^\star \mid q, a_{1:K}),
\end{equation}
where each question $q$ is presented with all $K$ answer options
$a_{1:K}$ and $y^\star$ is the correct answer label. Per-benchmark
settings are given in Table~\ref{tab:sft_config}.

\begin{table}[ht]
\centering
\caption{Per-benchmark SFT and DPO training configuration.}
\label{tab:sft_config}
\resizebox{\columnwidth}{!}{
\begin{tabular}{lrrrr}
\toprule
\textbf{Benchmark} & \textbf{Train} & \textbf{SFT Steps} & \textbf{Batch} & \textbf{DPO Steps} \\
\midrule
MedMCQA    & 182{,}822 & 8{,}000 & 32 & 2{,}000 \\
MedQA & 10{,}178  & 4{,}000 & 8  & 500 \\
PubMedQA   & 211{,}269 & 6{,}000 & 64 & 2{,}000 \\
\bottomrule
\end{tabular}
}
\smallskip
\parbox{\columnwidth}{\scriptsize
All runs: LR $2\times10^{-5}$ (SFT); LR $5\times10^{-6}$, $\beta=0.1$ (DPO).\\
MMLU has no training set; it is evaluated with the MedMCQA model.
}
\end{table}

\subsubsection{Direct Preference Optimization (DPO).}
Each SFT model serves as the reference policy $p_{\text{ref}}$ for its
DPO run. We build preference pairs per benchmark by pairing the correct
answer (preferred) with each incorrect answer (dispreferred),
$(q, y^+, y^-)$, and minimize the DPO
objective~\cite{rafailov2024direct}
\begin{equation}\label{eq:dpo}
\mathcal{L}_{\text{DPO}}
= -\,\mathbb{E}_{(q,y^+,y^-)}
\log \sigma\!\left(
\beta \log \tfrac{p_\theta(y^+\mid q)}{p_{\text{ref}}(y^+\mid q)}
- \beta \log \tfrac{p_\theta(y^-\mid q)}{p_{\text{ref}}(y^-\mid q)}
\right),
\end{equation}
at learning rate $5 \times 10^{-6}$ with temperature $\beta = 0.1$. The
KL-penalty strength in DPO scales as $1/\beta$, so the small $\beta$
permits large deviation from the reference and amplifies the attained
preference margin; we quantify this amplification in the results. On
MedMCQA, validation loss decreases from 0.6699 (step 500) to 0.6611
(step 2{,}000), which we select as the best checkpoint.

\subsubsection{Inference.}
We score all MedLLM evaluations by top-token selection: for a
multiple-choice question $q$ with options $\{a_1, \ldots, a_K\}$ we
predict $\hat{a} = \arg\max_k \log p_\theta(a_k \mid q)$, which avoids
the generation and parsing artifacts that penalize small models. Both
finetuned MedLLM and the DFT-only pretrained model are evaluated
zero-shot. For 7B baselines we follow~\citet{chen2023meditron}, using
3-shot in-context learning in the pretrained comparison and top-token
selection after finetuning.

\section{Experiments}

\subsection{Datasets}
We evaluate on four standard medical benchmarks spanning two task types. PubMedQA~\cite{jin2019pubmedqa} is context-grounded: each item pairs a PubMed abstract with a yes/no/maybe question, and we use the 500 expert-labeled examples in the reasoning-required setting. The remaining three are knowledge-recall benchmarks. MedMCQA~\cite{pal2022medmcqa} contains over 194K 4-option questions from Indian medical entrance exams (AIIMS/NEET) across 2{,}400 topics and 21 subjects; we evaluate on the 4{,}183-question validation set, since the test answer keys are not public. MedQA~\cite{jin2021disease} consists of USMLE-style clinical vignettes; we use the 4-option English variant, training on \texttt{GBaker/MedQA-USMLE-4-options} (10{,}178 examples) and evaluating on the 1{,}273-question test set (MedQA). MMLU~\cite{hendrycks2021measuring} aggregates nine medical subjects (1{,}862 questions) and provides no training set; we evaluate the MedMCQA-finetuned model on it to measure cross-domain transfer.

\subsection{Baselines}
We evaluate MedLLM against models from
$0.1$B to 7B: the medically adapted PMC-LLaMA-7B~\cite{wu2023pmcllama}, general-purpose
Falcon-7B~\cite{almazrouei2023falcon}, and instruction-tuned Mistral-7B-instruct~\cite{jiang2023mistral}
and Zephyr-7B-$\beta$~\cite{tunstall2023zephyr} (DPO-aligned), together
with the sub-7B open baselines Gemma-2B~\cite{gemmateam2024gemma},
BioMedLM-2.7B~\cite{bolton2024biomedlm}, and
BioGPT-Large~\cite{luo2022biogpt}. To match MedLLM's per-benchmark
fine-tuning, PMC-LLaMA-7B and Falcon-7B are individually fine-tuned on
each benchmark's training set, whereas the instruction-tuned Mistral-7B
and Zephyr-7B-$\beta$ are evaluated with their released weights, since
they are designed for zero-shot use.

\subsection{Experimental Settings}
\paragraph{Evaluation Protocol.}
We score MedLLM and all finetuned models by top-token selection, choosing the option with the highest log-probability among the answer candidates; MedLLM is evaluated zero-shot. Pretrained (non-finetuned) baselines use 3-shot in-context learning with demonstrations sampled from the training set across three random seeds. Because MMLU has no training set, its MedLLM result comes from the MedMCQA-finetuned model and is a cross-domain transfer measurement rather than in-domain performance.

\paragraph{Training Configuration.}
General pretraining runs on 2$\times$ NVIDIA H100 GPUs; SFT and DPO run on 2$\times$ NVIDIA A100 GPUs. Phase~A uses 200K steps on 29.5B tokens with curriculum lengths $128\to256\to512$; DFT uses 25K steps at LR $5\times10^{-5}$, carrying the final checkpoint forward. Phase B (SFT, LR $2\times10^{-5}$) and Phase C (DPO, LR $5\times10^{-6}$, $\beta=0.1$) are run per benchmark, with settings in Table~\ref{tab:sft_config}.

\section{Results}

\subsection{Pretrained Model Evaluation}
\begin{figure}[t]
\centering
\includegraphics[width=\columnwidth]{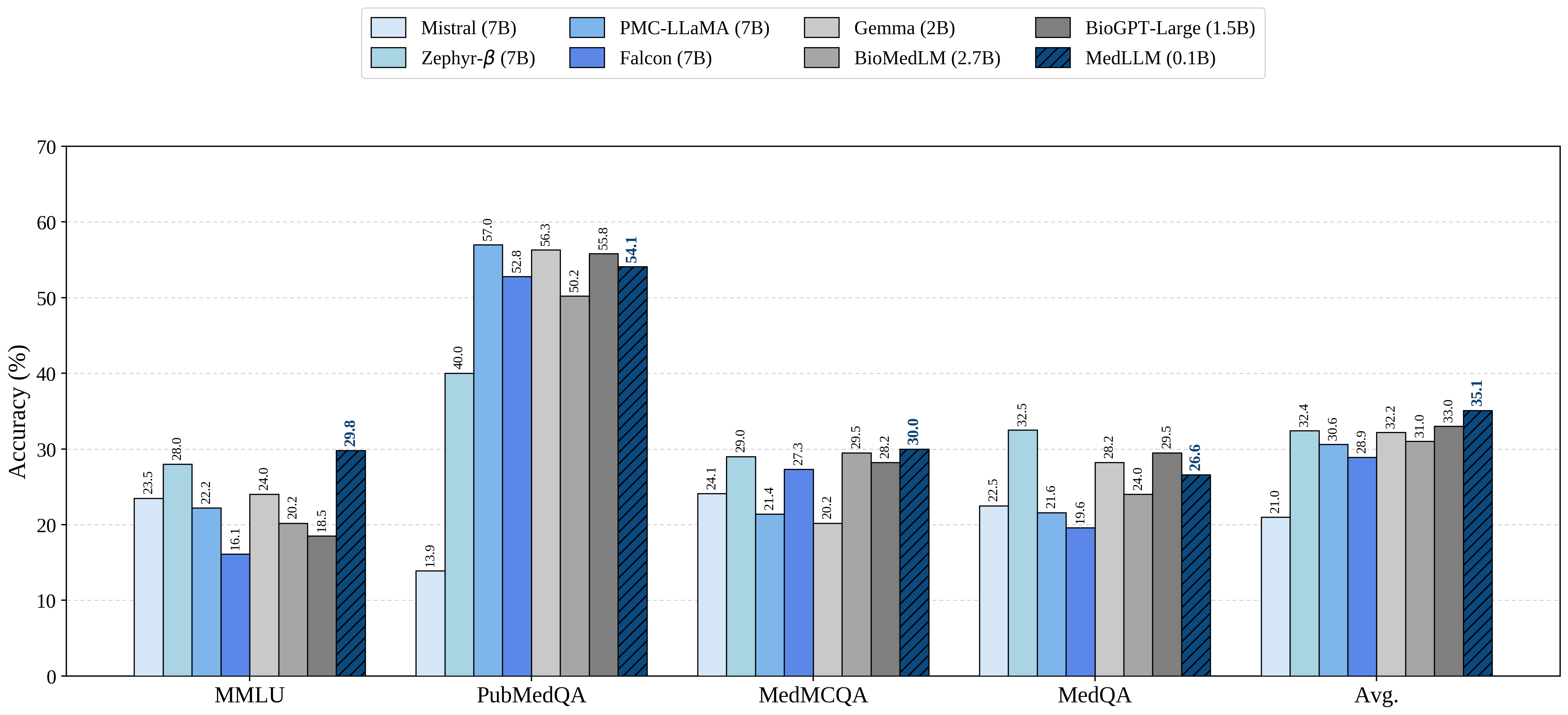}
\caption{Pretrained model comparison across scales.}
\label{fig:pretrained_comparison}
\end{figure}

Figure~\ref{fig:pretrained_comparison} shows results for the pretraining, with MedLLM evaluated after DFT only.
The comparison is conservative for MedLLM, since the baselines use 3-shot
in-context learning whereas MedLLM is evaluated zero-shot by top-token
selection, in-context learning being unreliable at 0.1B parameters.
Despite this handicap and its $70\times$ smaller size, MedLLM attains the
highest average accuracy in the figure at $35.1\%$, ahead of every 7B and
sub-7B baseline including the instruction-tuned Zephyr-7B-$\beta$
($32.4\%$) and the medically adapted PMC-LLaMA-7B ($30.6\%$); the lead is
carried by the two recall-oriented multiple-choice benchmarks, where
MedLLM is the top model on both MMLU ($29.8\%$, ahead of Zephyr-7B-$\beta$
at $28.0\%$) and MedMCQA ($30.0\%$, ahead of BioMedLM-2.7B at $29.5\%$),
showing that domain adaptation on MedFineWeb already lifts the small model
above larger systems on these tasks before any task-specific finetuning. The pretrained comparison already exhibits the split that finetuning later sharpens: the 0.1B model
is competitive-to-leading where the answer is a short factual or
context-grounded choice, and weakest on the benchmark demanding
multi-step clinical recall.

\subsection{Fine-Tuned Model Evaluation} 
Table~\ref{tab:finetuned_results} shows results after task-specific finetuning, and the ordering separates by task type. On the two multiple-choice recall benchmarks, MedLLM (DPO) outperforms, reaching $32.4\%$ on MMLU and $34.9\%$ on MedMCQA and significantly exceeding every 7B and sub-7B baseline under McNemar's test, including the medically adapted PMC-LLaMA-7B. On MedQA it does not clear this bar, at $28.1\%$ only $3.1$~pp above the four-option floor and significantly behind the instruction-tuned baselines, so its recall competence does not extend to longer USMLE vignettes. On context-grounded PubMedQA the ordering inverts: MedLLM reaches $58.2\%$ after DPO, significantly ahead of Zephyr-7B-$\beta$ ($43.0\%$) and Mistral-7B ($17.8\%$) but significantly behind the PubMedQA-finetuned PMC-LLaMA-7B ($72.9\%$) and Falcon-7B ($65.3\%$), so at $70\times$ fewer parameters the model clears the instruction-tuned systems on context while the continued-pretraining baselines retain a supervision advantage. Finally, DPO adds accuracy over SFT (MedMCQA $33.9\!\to\!34.9\%$, MedQA $+1.5$~pp, PubMedQA $+0.3$~pp, a slight MMLU regression) yet amplifies the mean correct-versus-incorrect log-probability margin $5.2\times$, from $0.1037$ to $0.5403$, the $O(1/\beta)$ separation the objective predicts at $\beta=0.1$ (Eq.~\ref{eq:dpo}), showing that at this scale preference alignment sharpens discrimination the model already has rather than supplying absent knowledge, consistent with stored knowledge being the binding constraint at 0.1B.

\begin{table*}[ht]
\centering
\caption{Task-specific finetuning results, showing acc $\pm$ std. over three seeds. Baselines marked with $^*$ are
instruction-tuned. Significance is assessed per benchmark by McNemar's
test, comparing the item-level predictions
of MedLLM (DPO) against each baseline: $\dagger$ marks a baseline that
MedLLM (DPO) significantly exceeds, and $\ddagger$ marks a baseline by
which MedLLM (DPO) is significantly exceeded ($p<0.05$).}
\label{tab:finetuned_results}
\resizebox{0.75\textwidth}{!}{
\begin{tabular}{llccccc}
\toprule
\textbf{Model} & \textbf{Size} & \textbf{MMLU} & \textbf{PubMedQA} & \textbf{MedMCQA} & \textbf{MedQA} & \textbf{Avg.} \\
\midrule
\multicolumn{7}{l}{\textit{7B-scale (top-token selection)}} \\
Mistral$^*$        & 7B & $25.8 \pm 0.9^{\dagger}$ & $17.8 \pm 1.5^{\dagger}$ & $30.2 \pm 0.6^{\dagger}$ & $41.1 \pm 1.1^{\ddagger}$ & 28.7 \\
Zephyr-$\beta$$^*$ & 7B & $32.1 \pm 1.0$ & $43.0 \pm 1.8^{\dagger}$ & $33.0 \pm 0.6^{\dagger}$ & $\textbf{42.5} \pm 1.2^{\ddagger}$ & 37.7 \\
PMC-LLaMA          & 7B & $26.4 \pm 0.9^{\dagger}$ & $\textbf{72.9} \pm 1.6^{\ddagger}$ & $26.6 \pm 0.6^{\dagger}$ & $25.5 \pm 1.0^{\dagger}$ & 37.9 \\
Falcon             & 7B & $23.6 \pm 0.8^{\dagger}$ & $65.3 \pm 1.7^{\ddagger}$ & $31.8 \pm 0.6^{\dagger}$ & $21.0 \pm 0.9^{\dagger}$ & 35.4 \\
\midrule
\multicolumn{7}{l}{\textit{sub-7B-scale (top-token selection)}} \\
Gemma        & 2B    & $27.6 \pm 0.9^{\dagger}$ & $60.8 \pm 1.8^{\ddagger}$ & $27.7 \pm 0.6^{\dagger}$ & $30.6 \pm 1.0^{\dagger}$ & 36.7 \\
BioMedLM     & 2.7B  & $23.8 \pm 0.8^{\dagger}$ & $55.2 \pm 1.8$ & $32.2 \pm 0.6^{\dagger}$ & $30.6 \pm 1.0^{\dagger}$ & 35.5 \\
BioGPT-Large & 1.5B  & $21.6 \pm 0.8^{\dagger}$ & $60.1 \pm 1.8^{\ddagger}$ & $33.1 \pm 0.6^{\dagger}$ & $31.3 \pm 1.0^{\dagger}$ & 36.5 \\
\midrule
\multicolumn{7}{l}{\textit{MedLLM (0.1B, top-token selection)}} \\
MedLLM (SFT) & 0.1B & $29.7 \pm 1.0$ & $57.9 \pm 1.8$ & $33.9 \pm 0.6$ & $26.6 \pm 1.0$ & 37.0 \\
MedLLM (DPO) & 0.1B & $\textbf{32.4} \pm 1.0$ & $58.2 \pm 1.8$ & $\textbf{34.9} \pm 0.6$ & $28.1 \pm 1.0$ & \textbf{38.4} \\
\bottomrule
\end{tabular}
}
\end{table*}

\begin{table*}[ht]
\centering
\caption{Ablation on training phases evaluated by top-token selection on
the per-benchmark fine-tuned models.}
\label{tab:ablation}
\resizebox{0.65\textwidth}{!}{
\begin{tabular}{lccccc}
\toprule
\textbf{Configuration} & \textbf{MMLU} & \textbf{PubMedQA} & \textbf{MedMCQA} & \textbf{MedQA} & \textbf{Avg.} \\
\midrule
SFT (\textit{w/o} DFT) & 31.1 & 55.0 & 32.8 & 25.8 & 36.2 \\
SFT (DFT Step 8k)      & 30.5 & 55.7 & 32.8 & 27.8 & 36.7 \\
SFT (DFT Step 20k)     & 30.4 & 55.5 & 33.6 & 27.3 & 36.7 \\
SFT (DFT Step 25k)     & 29.7 & 57.9 & 33.9 & 26.6 & 37.0 \\
SFT + DPO              & 32.4 & 58.2 & 34.9 & 28.1 & 38.4 \\
\bottomrule
\end{tabular}
}
\end{table*}

\subsection{Analysis on Continued Pretraining}
\begin{figure}[t]
\centering
\includegraphics[width=0.9\columnwidth]{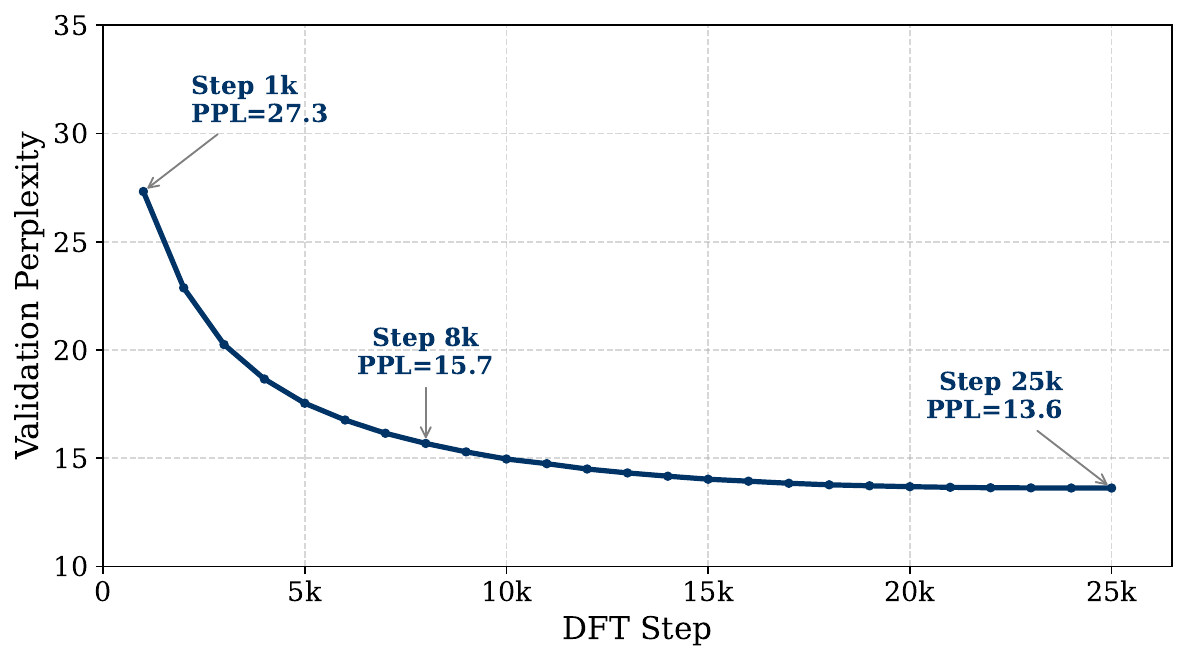}
\caption{Validation perplexity during medical domain fine-tuning.}
\label{fig:dft_loss}
\end{figure}

\paragraph{Training Dynamics.}
Figure~\ref{fig:dft_loss} shows validation perplexity during domain
fine-tuning. It decreases monotonically, from $27.3$ at step $1{,}000$ to
$15.68$ at step $8{,}000$ and $13.62$ at step $25{,}000$, with no
overfitting at any checkpoint, so we carry the final checkpoint into all
SFT and DPO experiments. The curve is strongly front-loaded: perplexity
halves within the first $10{,}000$ steps and improves by only about $1.3$
over the remaining $15{,}000$, indicating that a 0.1B model absorbs the
distributional characteristics of the corpus quickly and then saturates. Perplexity on medical text falls
by half, yet MedMCQA accuracy after DFT reaches only $27.9\%$, barely
above the $25\%$ four-option floor. Fitting the medical token
distribution and acquiring retrievable medical facts are therefore
separable at this scale, consistent with the task-structured gap observed
after finetuning: domain adaptation reliably teaches the model medical
language, but not the stored knowledge that recall benchmarks require.

\paragraph{Intermediate Checkpoint Evaluation.}
Table~\ref{tab:intermediate} makes this separation explicit by tracking
downstream accuracy across the same fine-tuning run, evaluated few-shot
following~\citet{chen2023meditron}. Over $1.26$B tokens of domain
fine-tuning, during which validation perplexity roughly halves, benchmark
accuracy does not move: MMLU shifts by $+0.4$~pp ($29.3\!\to\!29.7$),
MedMCQA by $+0.2$~pp ($26.5\!\to\!26.7$), PubMedQA by $-0.3$~pp
($33.8\!\to\!33.5$), and the average stays at $\approx\!30\%$
($29.9\!\to\!30.0$), all within seed noise. The two curves therefore
diverge: perplexity improves substantially while accuracy is flat, which
is direct evidence that at 0.1B the corpus teaches medical form (lower
perplexity) without depositing the retrievable facts that these
benchmarks score. It also implies that perplexity is an unreliable
model-selection signal at this scale, since the lowest-perplexity
checkpoint is not the most accurate one.

\begin{table}[ht]
\centering
\caption{Few-shot evaluation of intermediate DFT checkpoints.}
\label{tab:intermediate}
\resizebox{\columnwidth}{!}{
\begin{tabular}{lccccc}
\toprule
\textbf{Step} & \textbf{Tokens} & \textbf{MMLU} & \textbf{PubMedQA} & \textbf{MedMCQA} & \textbf{Avg.} \\
\midrule
0 (base) & 0 & 29.3 & 33.8 & 26.5 & 29.9 \\
2,000 & ${\sim}$126M & 27.7 & 33.8 & 26.6 & 29.4 \\
5,000 & ${\sim}$315M & 28.3 & 33.7 & 26.8 & 29.6 \\
8,000 & ${\sim}$504M & 29.2 & 33.5 & 27.1 & 29.9 \\
20,000 & ${\sim}$1.26B & 29.7 & 33.5 & 26.7 & 30.0 \\
\bottomrule
\end{tabular}
}
\end{table}

\subsection{Ablation Studies}
Table~\ref{tab:ablation} isolates each training phase, evaluated on the
per-benchmark fine-tuned models. Domain fine-tuning alone barely moves
aggregate accuracy, which rises modestly from $36.2$ to $37.0$ across $25{,}000$ steps
($36.2\!\to\!36.7\!\to\!36.7\!\to\!37.0$) even as validation perplexity
halves over the same interval ($27.3\!\to\!13.62$,
Figure~\ref{fig:dft_loss}); what DFT changes is the distribution of
competence, not its level, with MedMCQA rising ($32.8\!\to\!33.9$) and
MMLU declining ($31.1\!\to\!29.7$), while MedQA remains roughly stable ($25.8\!\to\!26.6$), the
signature of capacity reallocation rather than knowledge acquisition at
0.1B. DPO then adds a genuine \(+1.4\) pp to the average (\(37.0 \to 38.4\)), but concentrated rather than broad: it gains \(+1.0\) pp on MedMCQA and \(+2.7\) pp on MMLU transfer while leaving context-grounded PubMedQA flat (\(57.9 \to 58.2\)) and recovering MedQA to \(28.1\). The same step
amplifies the mean log-probability margin between correct and incorrect
answers $5.2\times$, from $0.1037$ to $0.5403$, the $O(1/\beta)$
separation the objective predicts at $\beta=0.1$ (Eq.~\ref{eq:dpo}), so
preference alignment sharpens discrimination the model already possesses
far more than it supplies new knowledge, and aggregate accuracy understates both phases'
effects, which are task-structured redistributions the average partly cancels.

\section{Conclusions}
We presented MedLLM, an open 0.1B-parameter medical language model, and MedFineWeb, the web-selected corpus on which it is domain adapted, both released to make sub-billion medical modeling reproducible. Against 7B and sub-7B baselines, medical competence does not degrade uniformly under compression but splits by task type. On context-grounded PubMedQA the model falls within $2.9$~pp of a medically adapted 7B system at $70\times$ fewer parameters, and on the recall benchmark MedMCQA it significantly exceeds every 7B and sub-7B baseline, so where recall succeeds the binding constraint is model capacity rather than adaptation; on the harder USMLE-style MedQA, however, it stays near the four-option floor, the one task on which the capacity limit is not overcome. At sub-billion scale, medical capability is therefore best sourced from retrieval and supplied context rather than from parametric recall. MedLLM remains constrained in the breadth of knowledge it can encode and our log-probability ranking does not assess free-form generation.

\bibliography{aaai2027}

\end{document}